\documentclass[letterpaper, 10 pt, conference]{ieeeconf}
\IEEEoverridecommandlockouts    

\usepackage{multirow}
\usepackage{lipsum}
\usepackage{amsmath}
\usepackage{amssymb}
\usepackage{graphicx}
\graphicspath{{./Figures/}}
\usepackage{float}

\usepackage{algorithm}
\usepackage{algorithmicx}
\usepackage{algpseudocode}
\usepackage{booktabs}
\title{\LARGE \bf Real2Sim: A Physics-driven and Editable Gaussian Splatting Framework for Autonomous Driving Scenes}

\author{
	\parbox{\textwidth}{%
		\centering
		Kaicong Huang$^{1}$, Talha Azfar$^{1}$, Weisong Shi$^{2}$, Ruimin Ke$^{1\star}$%
	}%
	\thanks{$^{1}$ Department of Civil and Environmental Engineering, Rensselaer Polytechnic Institute, 110 Eighth Street, Troy, NY USA 12180.
		{\tt\small huangk10@rpi.edu, azfart@rpi.edu, ker@rpi.edu}}%
	\thanks{$^{2}$ Department of Computer and Information Sciences, University of Delaware, Newark, DE USA 19716.
		{\tt\small weisong@udel.edu}}%
	\thanks{$^{\star}$ Corresponding author.}%
}

\begin{document}
	
\maketitle
\thispagestyle{empty}
\pagestyle{empty}

\begin{abstract}
Reliable autonomous driving relies on large-scale, well-labeled data and robust models. However, manual data collection is resource-intensive, and traditional simulation suffers from a persistent reality gap. While recent generative frameworks and radiance-field methods improve visual fidelity, they still struggle with temporal and spatial consistency and cannot ensure physics-aware behavior, limiting their applicability to driving scenario generation. To address these challenges, we propose Real2Sim, an unified framework that combines 4D Gaussian Splatting (4DGS) with a differentiable Material Point Method (MPM) solver. Real2Sim explicitly reconstructs dynamic driving scenes as temporally continuous Gaussian primitives, supports instance-level editing, and simulates realistic object-object and object-environment interactions. This framework enables physics-aware, high-fidelity synthesis of diverse, editable scenarios, including challenging corner cases such as collisions and post-impact trajectories. Experiments on the Waymo Open Dataset validate Real2Sim’s capabilities in rendering, reconstruction, editing, and physics simulation, demonstrating its potential as a scalable tool for data generation in downstream tasks such as perception, tracking, trajectory prediction, and end-to-end policy learning.
\end{abstract}

\begin{figure}[t]
  \centering
  \includegraphics[width=1.0\columnwidth]{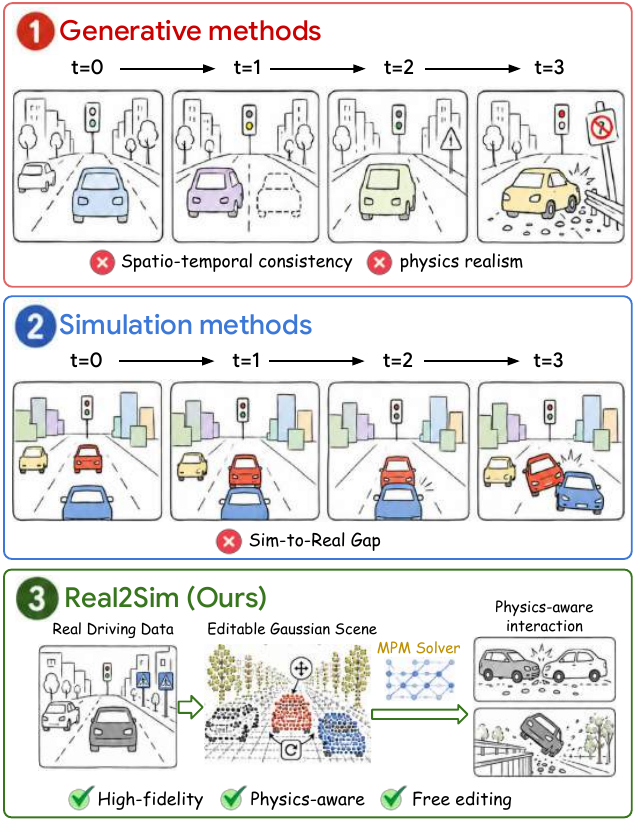}
  \caption{Comparison of scene generation paradigms for autonomous driving. Generative methods often lack spatio-temporal consistency and physical realism, while conventional simulation methods suffer from a sim-to-real gap. Real2Sim bridges these limitations by integrating Gaussian Splatting with differentiable physics solver.}
  \label{fig:intro}
\end{figure}

\begin{figure*}[t]
  \centering
  \includegraphics[width=0.9\textwidth]{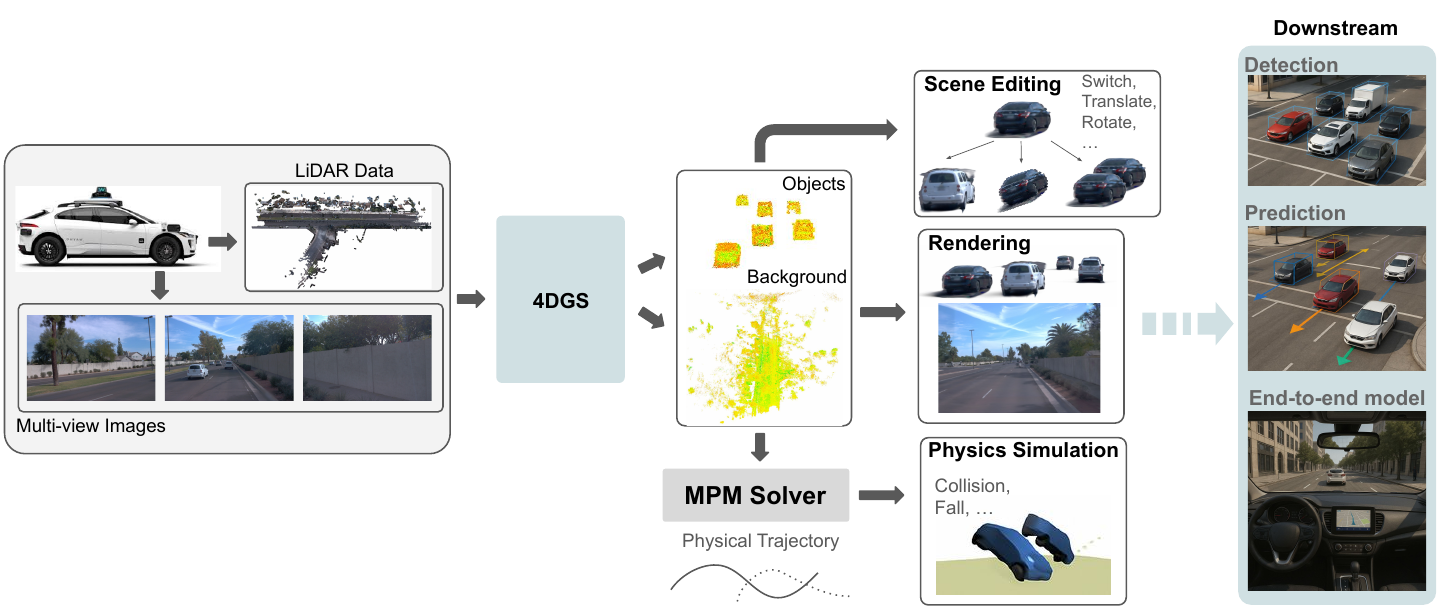}
  \caption{Framework of Real2Sim. Real2Sim first reconstructs separate Gaussian-based representations of the static background and each dynamic object from multi-modal input data. Each object is encoded as an independent Gaussian cluster, enabling high-degree-of-freedom editing to create diverse traffic scenarios. To impart physical behavior, a lightweight physics engine is introduced, treating each Gaussian as a particle whose state is updated by the MPM solver. The resulting synthetic data can support downstream tasks such as detection, tracking, motion prediction, and end-to-end driving model training.}
  \label{fig:8}
\end{figure*}
	
\section{Introduction}
Reliable autonomous driving systems hinge on two pillars: powerful, robust models and large‑scale, well‑labeled data. Recent foundation driving models report performance gains when trained on millions of time‑synchronized images or video frames with dense 2D/3D labels \cite{sun2020scalability}, an effort that can require tens of thousands of staff‑hours using conventional manual pipelines.

To ease this burden, researchers have turned to data‑generating simulators and generative models. Methods such as SurfelGAN \cite{yang2020surfelgan} synthesize camera and LiDAR signals directly from limited drives, and the most recent work, GAIA‑2 \cite{russell2025gaia}, realizes multi‑view generation and multi‑agent interactions through a diffusion world model. Such works shrink the gap between real and synthetic sensors, yet they still suffer from rendering artifacts and a noticeable domain gap when models trained in simulation are deployed in the real world. And aligning multi‑view outputs and keeping geometry consistent over time remain open challenges.

In recent years, Neural Radiance Field (NeRF) \cite{mildenhall2021nerf} pushed high‑fidelity image generation forward, but their volumetric field evaluation is too slow for closed‑loop simulation. The advent of 3D Gaussian Splatting (3DGS) \cite{kerbl20233d} achieves comparable quality while enabling real‑time rendering of whole street blocks on commodity GPUs, making it attractive for robotics and autonomous driving research. Building on this, 4D Gaussian Splatting (4DGS) \cite{wu20244d, duan20244d, zhou2024drivinggaussian, yan2024street} extend the representation with temporal deformation, supporting continuous novel‑time, novel‑view synthesis for complex scenes with moving vehicles and pedestrians.

Nevertheless, existing 4DGS pipelines record what the sensors saw and edit the current Gaussian primitives naively (position, orientation, mesh replacement). They cannot simulate what could happen in the future under the simulation settings. In particular, they lack physics‑aware behavior, so edited objects do not react plausibly to forces or collisions. These capabilities are indispensable for testing corner cases such as multi‑car crashes or near‑misses in safety‑critical domains. Recent efforts to embed kinematics and dynamics into Gaussian primitives \cite{xie2024physgaussian, borycki2024gasp, jiang2024vr} demonstrate feasibility but have not yet been explored in street-wide driving sequences.

To enable a high‑fidelity, editable, and physically embedded framework for autonomous driving simulation and scene generation, we introduce Real2Sim, a system that unifies 4D Gaussian Splatting with a differentiable physics solver to turn real‑world driving scenes into editable, physics‑aware simulations, as shown in Fig. \ref{fig:intro}. Real2Sim first reconstructs a continuous 4D Gaussian representation of each scene clip, capturing appearance and motion over the entire time window. It then assigns every Gaussian cluster to an object instance so that users can reposition, rotate, scale, or even replace vehicles, pedestrians, and static assets on the fly. Each object carries mass, friction, and restitution parameters, and by integrating these properties with a Material Point Method (MPM) solver, the system updates all Gaussians physically and reproduces realistic object-object and object-environment interactions, including collisions and post‑impact trajectories. 
Such editing with high degree of freedom combined with physical simulation not only benefits the generation of diverse driving scenarios but also addresses the lack of corner case data in autonomous driving datasets, which often results in long-tailed training distributions \cite{zhang2023deep}.
We test Real2Sim on the Waymo Open Dataset \cite{sun2020scalability} and demonstrate dynamic reconstruction, instance‑level editing, and physical simulation to support abundant and diverse training data for tasks such as object detection, multi‑object tracking, trajectory prediction, and end‑to‑end driving policy learning.

The main contributions of this work could be summarized as follows:
\begin{itemize}
\item We propose Real2Sim, a simulation framework for autonomous driving based on Gaussian Splatting that delivers high-fidelity, physics-aware scene generation and supports arbitrary in-scene editing, enabling the creation of diverse, real-world traffic scenarios.

\item Real2Sim allows users to customize the physical properties of every object and faithfully simulates object‑to‑object and object‑to‑environment interactions, increasing the flexibility and realism of synthesized traffic scenes.

\item Real2Sim supports the creation of corner cases and large-scale synthetic datasets that benefit downstream tasks such as detection, tracking, motion prediction, and end-to-end driving.

\end{itemize}

\section{Related Works}
\subsection{3D Vision for Scene Reconstruction and Autonomous Driving}
3D vision can accurately reconstruct the spatial relationships of objects in the physical world, and has therefore attracted widespread attention in autonomous driving and robotics. Early reconstruction methods relied primarily on multi‐view stereo (MVS) and structure‐from‐motion (SfM) to recover dense point clouds from multiple calibrated cameras. A representative method, COLMAP, generates high‐quality geometry through offline optimization over large image collections. However, it is computationally intensive and performs poorly on dynamic objects \cite{schonberger2016structure}.

With the advent of deep learning, point‐based and voxel‐based networks have been widely applied to construct 3D structure directly from images or LiDAR scans \cite{qi2017pointnet, zhou2018voxelnet, choy20194d}. More recently, neural implicit representations such as NeRF encode a scene as a continuous volumetric function and use a small MLP to synthesize high‐fidelity novel views \cite{mildenhall2021nerf}. Subsequent work has focused on integrating various downstream tasks and accelerating rendering \cite{li2021mine, kundu2022panoptic}. Although these methods deliver excellent visual quality, they require dense per‐scene optimization, making them impractical for real‐time autonomous driving applications \cite{gao2022nerf}.

Gaussian Splatting instead replaces implicit voxels with explicit Gaussian kernels \cite{kerbl20233d}, enabling real‐time scene editing and demonstrating exceptional performance in both indoor and urban outdoor reconstruction. This approach supports diverse data generation for downstream tasks such as detection, segmentation, tracking, prediction, and end‐to‐end autonomous driving models \cite{zhu20243d}. The emergence of 4D Gaussian representations, which directly model dynamic objects and scenes, makes them particularly well-suited for motion-rich driving environments \cite{wu20244d, duan20244d, zhou2024drivinggaussian, yan2024street}.

In the autonomous driving domain, large‐scale multimodal datasets such as the Waymo Open Dataset \cite{sun2020scalability} and nuScenes \cite{caesar2020nuscenes} provide extensive sequences for perception and mapping research. Simulation platforms like CARLA \cite{dosovitskiy2017carla} have become essential for testing perception and planning algorithms. However, according to the scaling laws of large‐model training, vast amounts of data remain critical for all major autonomous‐driving developers. 
Collecting real-world data is both time-consuming and costly, and conventional simulations frequently exhibit a notable reality gap. Gaussian Splatting overcomes these limitations by enabling efficient, laboratory-scale data editing and synthesis. When paired with real-time, physics-based scene simulation, it not only delivers accurate rendering and reconstruction but also supports dynamic, in-scene interactions. This unified platform streamlines both simulation and data generation for downstream tasks such as detection, tracking, motion prediction, and end-to-end driving model training.

Building on these insights, we introduce Real2Sim: a comprehensive autonomous-driving simulation platform that combines Gaussian Splatting’s high-fidelity rendering and interactive scene editing with Gaussian‐level physics simulation, delivering rich, versatile data for downstream tasks.

\section{Methodology}
We propose Real2Sim, a framework that integrates Gaussian Splatting with an MPM solver to generate and edit real-world driving scenes with physics awareness, as shown in Fig.~\ref{fig:8}. Real2Sim first reconstructs separate Gaussian-based representations of the static background and each dynamic object in a video sequence (Section 3.B). Each object is encoded as an independent Gaussian cluster, allowing users to reposition or rotate vehicles or to insert external 3D assets to diversify traffic scenarios. To impart physical behavior, Real2Sim incorporates a lightweight physics engine that treats each Gaussian as a particle whose state is updated by the MPM solver (Section 3.C). This approach faithfully reproduces object-object and object-environment interactions, including collisions and subsequent trajectories. The remainder of this section first reviews the fundamentals of 3D Gaussian Splatting and then describes Real2Sim’s core modules in detail.

\subsection{3D Gaussian Splatting}
3D Gaussian Splatting (3DGS) \cite{kerbl20233d} represents a static scene as an explicit collection of anisotropic Gaussian kernels, each parameterized by a center position, covariance, opacity, and spherical‐harmonic color coefficients. Formally, each Gaussian \(G_p\) at world‐space position \(\mathbf{x}_p\) is defined by its covariance matrix \(\Sigma_p\) and mean \(\mathbf{x}_p\):
\begin{equation}
G_p(\mathbf{x}) \;=\; \exp\Bigl(-\tfrac{1}{2}(\mathbf{x}-\mathbf{x}_p)^\top \Sigma_p^{-1} (\mathbf{x}-\mathbf{x}_p)\Bigr)
\end{equation}

The covariance is decomposed as:
\begin{equation}
\Sigma_p = R_p\,S_p\,S_p^\top\,R_p^\top
\end{equation}
where \(R_p\) is a rotation and \(S_p\) is a diagonal scaling matrix.

To render a novel view, each 3D Gaussian is projected into the image plane via differentiable splatting \cite{yifan2019differentiable}. Given a camera‐to‐world affine Jacobian \(J\) and a viewing transform matrix \(W\), the covariance in camera coordinates becomes:
\begin{equation}
\Sigma_p' = JW\Sigma_pWJ^\top
\end{equation}

And the resulting 2D Gaussians contribute to pixel color \(C(\mathbf{u})\) through front‐to‐back compositing:
\begin{equation}
C(\mathbf{u}) \;=\; \sum_{p=1}^N \alpha_p(\mathbf{u})\,\mathbf{c}_p \;\prod_{q<p}\bigl(1-\alpha_q(\mathbf{u})\bigr)
\end{equation}
where \(\alpha_p\) is the effective opacity and \(\mathbf{c}_p\) the color from spherical harmonics.

This explicit, point‐based formulation replaces costly volumetric integration with efficient kernel splatting, enabling real‐time rendering at high resolution and facilitating direct scene manipulation. In particular, 3DGS permits per‐object editing (e.g., translation, rotation, insertion of external models) and provides a natural interface for physics integration, as demonstrated in Section 3.3 where Gaussian kernels become material points in a MPM solver.

\subsection{Driving Scene Construction}
To achieve editable and physics-aware simulation, we require a scene representation that is both explicit and efficient for real-time rendering, and that allows objects and background to be processed separately over time. We adopt the Street Gaussians methodology~\cite{yan2024street}. This approach models a driving sequence as two interleaved Gaussian point clouds: one for the static background and one for each moving object.

The background Gaussians are initialized from a sparse LiDAR point set combined with structure-from-motion (SfM) outputs and then aggregated into a dense field of 3D Gaussians through training. Each primitive is defined by:
\begin{itemize}
  \item a mean position $\mu$,
  \item an anisotropic covariance matrix $\Sigma_{b} = R_{b}\,S_{b}\,S_{b}^\top\,R_{b}^\top$,
  \item an opacity weight $\alpha_{b}$, and
  \item view-dependent color coefficients $z_{b}$ encoded with low-order spherical harmonics.
\end{itemize}

Compared to voxel or mesh representations, this explicit form captures fine geometric detail with far fewer primitives, enabling both high-fidelity reconstruction and fast differentiable rendering.

For moving objects such as vehicles, an off-the-shelf detector is used to segment and track each instance. Unlike the background representation, each dynamic object is represented by a dedicated Gaussian cluster, initially defined in the object’s local frame and then transformed into world coordinates via learned object pose parameters $(R_{t}, T_{t})$. That is, each object’s translation and rotation can evolve over time.

Because object appearance varies continuously over time and space, standard spherical harmonics cannot capture instantaneous changes. To address this, a 4D Fourier spherical harmonics basis is introduced:
\begin{equation}
z_{m,\ell}(t) \;=\; \sum_{i=0}^{k-1} f_i \,\cos\biggl(\frac{i\pi}{N_t}\,t\biggr)
\end{equation}
where $z_{m,\ell}(t)$ is the spherical harmonics coefficient at time $t$ and $f_i$ are the Fourier coefficients. A small set of these coefficients reconstructs each object’s harmonics weights $z(t)$ on the fly, avoiding per-frame parameters while preserving view-dependent effects and temporal coherence.

At inference time, background and object point clouds are merged into a single point cloud and composite them front-to-back using their opacity weights $\alpha$. The explicit, instance-aware nature of this representation provides the foundation for editing and physics extensions. Because each object is represented by its own Gaussian cluster and pose, users can relocate or rotate vehicles arbitrarily to create new traffic scenarios. In the following section, we describe how we assign mass, friction, and restitution to these clusters and integrate a differentiable Material Point Method solver, transforming static reconstruction into a fully editable, physics-aware simulation environment.

\begin{algorithm}[t]
\caption{MPM Integration on Gaussian Primitives}
\label{alg:mpm_integration}
\begin{algorithmic}[1]
\Require Particles $\{p\}$ with $(m_p, V_p^0, x_p, v_p, F_p)$ \Comment{$m_p$: mass, $V_p^0$: rest volume, $x_p,v_p$: position/velocity, $F_p$: deformation gradient}
\Require Time step $\Delta t$ \Comment{simulation timestep}
\For{each timestep}
  \State \textbf{1) Particle-to-Grid (P2G):}
  \ForAll{grid nodes $i$}
    \State $m_i \gets \sum_{p} m_p\,w_{ip}$ \Comment{$w_{ip}$: interpolation weight}
    \State $(mv)_i \gets \sum_{p} m_p\,v_p\,w_{ip}$
  \EndFor

  \State \textbf{2) Compute grid forces:}
  \ForAll{grid nodes $i$}
    \State $P_p \gets \frac{\partial \Psi}{\partial F}(F_p)$ \Comment{$P_p$: first Piola-Kirchhoff stress, $\Psi$: hyperelastic energy}
    \State $f_i \gets -\sum_{p} V_p^0\,P_p\,\nabla w_{ip} + f_i^{\mathrm{ext}}$ \Comment{$f_i^{\mathrm{ext}}$: external forces}
  \EndFor

  \State \textbf{3) Update grid velocities:}
  \ForAll{grid nodes $i$}
    \State $v_i \gets \frac{(mv)_i + \Delta t\,f_i}{m_i}$
  \EndFor

  \State \textbf{4) Grid-to-Particle (G2P):}
  \ForAll{particles $p$}
    \State $v_p \gets \sum_{i} v_i\,w_{ip}$
    \State $x_p \gets x_p + \Delta t\,v_p$
    \State $F_p \gets \bigl(I + \Delta t\,\sum_{i} v_i\,\nabla w_{ip}^\top\bigr)\,F_p$ \Comment{$I$: identity matrix}
  \EndFor
\EndFor
\end{algorithmic}
\end{algorithm}

\subsection{Physics-aware Simulation}
To enable realistic physical interactions in edited driving scenes, Real2Sim integrates a lightweight Material Point Method (MPM) \cite{jiang2016material} solver directly with the Gaussian representation generated in the previous section. Each Gaussian primitive is treated as a material particle carrying mass $m_p$, initial volume $V_p^t$, position $x_p^t$, velocity $v_p^t$, and deformation gradient $F_p^t$. At each timestep, particle quantities are transferred to an Eulerian grid via $C^1$ B-spline kernels; the grid velocities are then updated, and the updated values are transferred back to particle velocities, positions, and deformation gradients. The overall algorithm for MPM integration is summarized in Algorithm \ref{alg:mpm_integration}.

To simulate the deformation of Gaussian primitives, following the settings in \cite{xie2024physgaussian}, we write down the first-order Fourier approximation of affine transformation as:
\begin{equation}
    \phi_p(X,t)=x_p(t)+F_p(t)\,(X-X_p)
\end{equation}

The Gaussian’s mean and covariance naturally evolve as  
\begin{equation}
x_p(t)=\phi(X_p,t),\quad A_p(t)=F_p(t)\,\Sigma_p\,F_p(t)^\top
\end{equation}
Then a static Gaussian with material‐space mean $X_p$ and covariance $A_p$ deforms to  
\begin{equation}
G_p(x,t)=\exp\Bigl(-\tfrac12\,(x-x_p(t))^\top A_p(t)^{-1} (x-x_p(t))\Bigr)
\end{equation}

Moreover, it is assumed that the opacity and color coefficients remain constant during simulation. The deformed Gaussian particles are then rendered with the original splatting pipeline.

By unifying 3D Gaussian Splatting with MPM, Real2Sim allows users to edit object poses and immediately simulate collisions, friction, and post‑impact trajectories without mesh extraction or remeshing, thus generating diverse, physics‑aware driving scenarios for downstream perception and control tasks.

\section{Experiments}
\subsection{Dataset}
We conduct our experiments on Waymo Open Dataset \cite{sun2020scalability}, which is a large-scale, high-fidelity benchmark for autonomous driving perception and motion tasks. Its Perception Dataset comprises 2,030 driving sequences, each 20s long and recorded at 10 Hz, yielding over 400,000 frames captured by five high-resolution (1920×1280) cameras and a 64‑beam LiDAR mounted on a production Waymo vehicle.  The original release included sequences across three U.S. cities: Phoenix, San Francisco, and Mountain View, annotated with 2D and 3D bounding boxes and consistent tracking IDs for 23 object classes.  

In our experiments, we utilize some sequences from the Perception Dataset for both reconstruction and downstream evaluation, leveraging its diverse scenarios and rich annotations to validate Real2Sim’s dynamic reconstruction, editing, and physics‑aware simulation capabilities.

\subsection{Implementation Details}
All training and inference are performed on a single RTX 4090 GPU with 24 GB of VRAM. For the StreetGaussian model, each scene is trained for 50,000 iterations using the same hyperparameters as the original paper. The physics simulation is decoupled from Gaussian model generation. Specifically, at each timestep, the GS point clouds of individual objects are loaded from the pretrained scene and evolved via MPM integration. The relevant physical parameters are detailed in the following section.

\subsection{Scene Rendering and Reconstruction}
Gaussian Splatting, due to its high-fidelity rendering capability and extremely fast throughput, has been widely adopted in autonomous driving and robotics. In this subsection we select two sequences from the Waymo Open Dataset for evaluation of scene rendering and reconstruction. Details are provided in Table \ref{tab:dataset}.

\begin{table}[h]
	\caption{Sequences and frame ranges used for scene rendering and reconstruction.}
  \label{tab:dataset}
	\begin{center}
		\begin{tabular}{c|cccc}
                \toprule
			Sequence ID & Cameras & Frame Range & Image Numbers & FPS \\
                \midrule
                002 & 5 & 50-100 & 505 & 28.57 \\
                031 & 3 & 0-198 & 594 & 28.72 \\
                \bottomrule
		\end{tabular}
	\end{center}
\end{table}

\begin{figure*}[!ht]
  \centering
  \begin{minipage}[t]{0.49\textwidth}
    \centering
    \includegraphics[width=\textwidth]{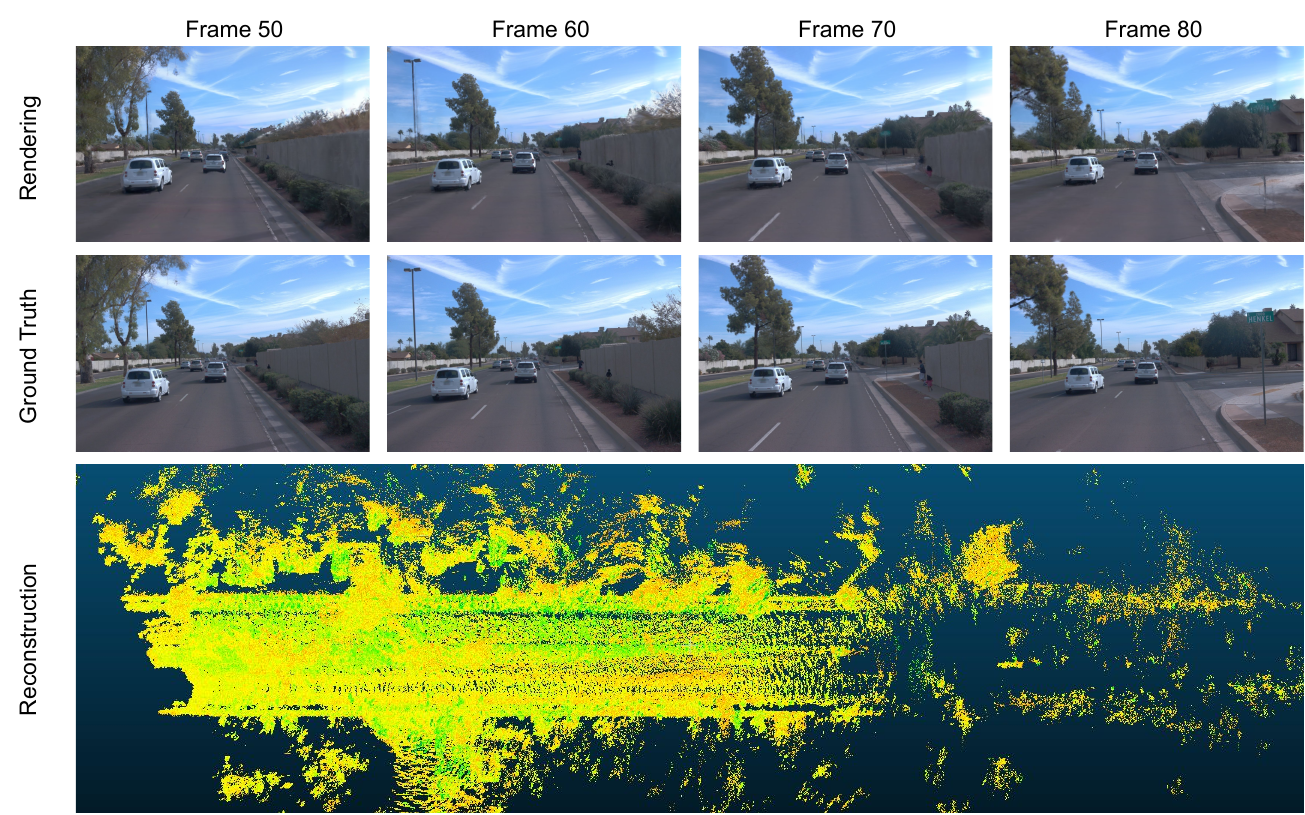}
    \caption{Rendering and reconstruction on Sequence 002. The renderer produces high-fidelity images while outputting Gaussian centers as points for point-cloud reconstruction.}
    \label{fig:1}
  \end{minipage}\hfill
  \begin{minipage}[t]{0.49\textwidth}
    \centering
    \includegraphics[width=\textwidth]{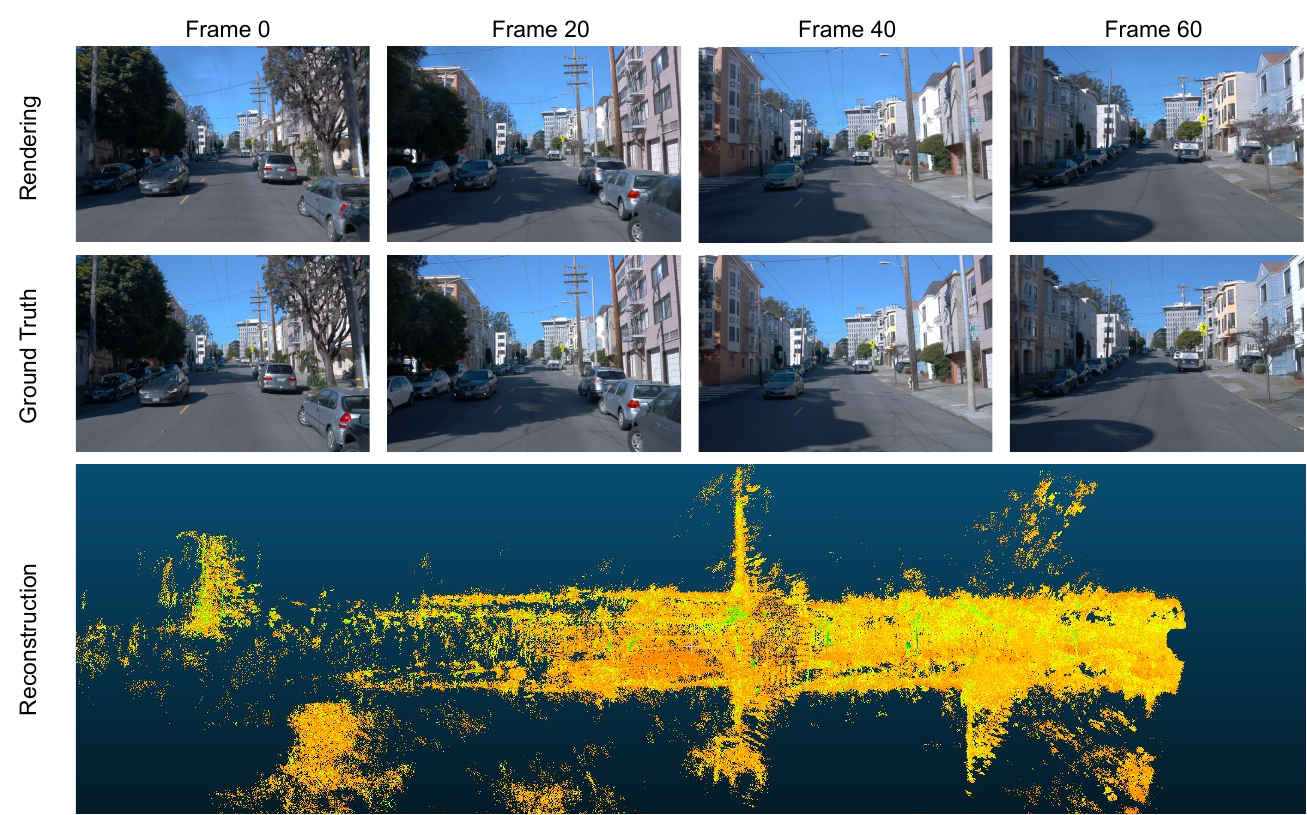}
    \caption{Rendering and reconstruction on Sequence 031. The scene is captured in a narrow street, and the renderer faithfully reproduces the shadows.}
    \label{fig:2}
  \end{minipage}
\end{figure*}

Fig. \ref{fig:1} presents the results on Sequence 002. We train on frames 50-100 using five cameras oriented to the left, front-left, front, front-right, and right. In Fig. \ref{fig:1}, renderings from the forward-facing camera for frames 50-80 are shown. These renderings closely match the ground truth. We also reconstruct the scene point cloud by taking the center of each Gaussian primitive as a point location, illustrating high-precision recovery of static elements such as the road surface, trees, and walls. Notably, dynamic objects and the static background are modeled separately, enabling object-level editing and providing a foundation for physics-based simulation.

Fig. \ref{fig:2} shows rendering and reconstruction on Sequence 031. Here, frames 0-198 from three cameras are used for training, and renderings for frames 0-60 are displayed. The model excels not only at capturing fine details of vehicles and buildings but also at accurately reproducing shadow effects.

\subsection{Scene Editing}
Gaussian Splatting method of explicitly constructing rendered scenes enables arbitrary, on-demand editing. Every object in the scene is represented by a cluster of Gaussian primitives, where each primitive’s rotation is a function of time $t$, and the object’s position and rotation both vary with $t$. Hence, by simply specifying an object’s position and rotation, one can effortlessly perform seamless in-scene modification.

\begin{figure}[t]
  \centering
  \includegraphics[width=1.0\columnwidth]{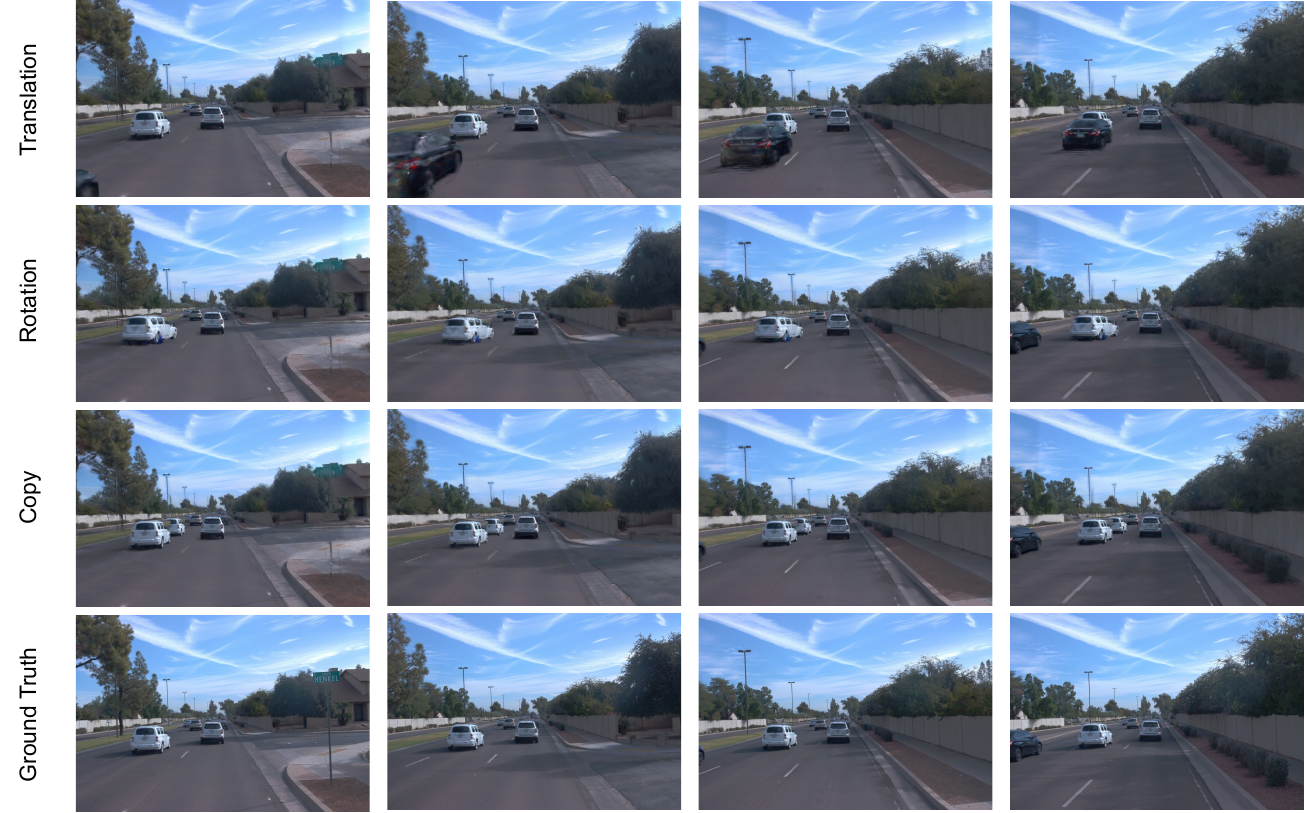}
  \caption{Scene editing examples: translation, rotation, and duplication. These results demonstrate the benefit of modeling each object separately and enabling arbitrary in-scene editing.}
  \label{fig:3}
\end{figure}

Fig. \ref{fig:3} presents three editing examples: translation, rotation, and duplication. In the first row, the black car is translated from the left lane to the center lane. In the second row, the white car in the center lane is rotated 15° to the right. In the third row, this vehicle is duplicated and then translated in the $x$-direction (i.e., the forward direction) by two car-lengths. These scene edits produce diverse road scenarios. Compared to introducing additional models from library, directly editing the existing scene physics yields more “realistic” and logically consistent data—scenarios that could plausibly occur in the same environment—thus providing rich training data for downstream tasks such as detection, tracking, and end-to-end autonomous driving models.

Our framework also supports loading and editing models external to the original scene. Although this may compromise realism, it offers significant benefits for data augmentation and model generalization, as demonstrated by the CRUISE system \cite{xu2025cruise}. While this capability is feasible, it is not explored further in the present work.

\subsection{Physics-aware Simulation and Corner Case Generation}
By explicitly endowing each Gaussian primitive with material properties and treating it as a particle, and by solving the dynamics via the Material Point Method (MPM), we achieve near-realistic physical simulation within Gaussian scenes. This approach captures both object-object and object-environment interactions such as collisions and post-impact trajectory generation.


\begin{figure*}[t]
  \centering
  \includegraphics[width=0.8\textwidth]{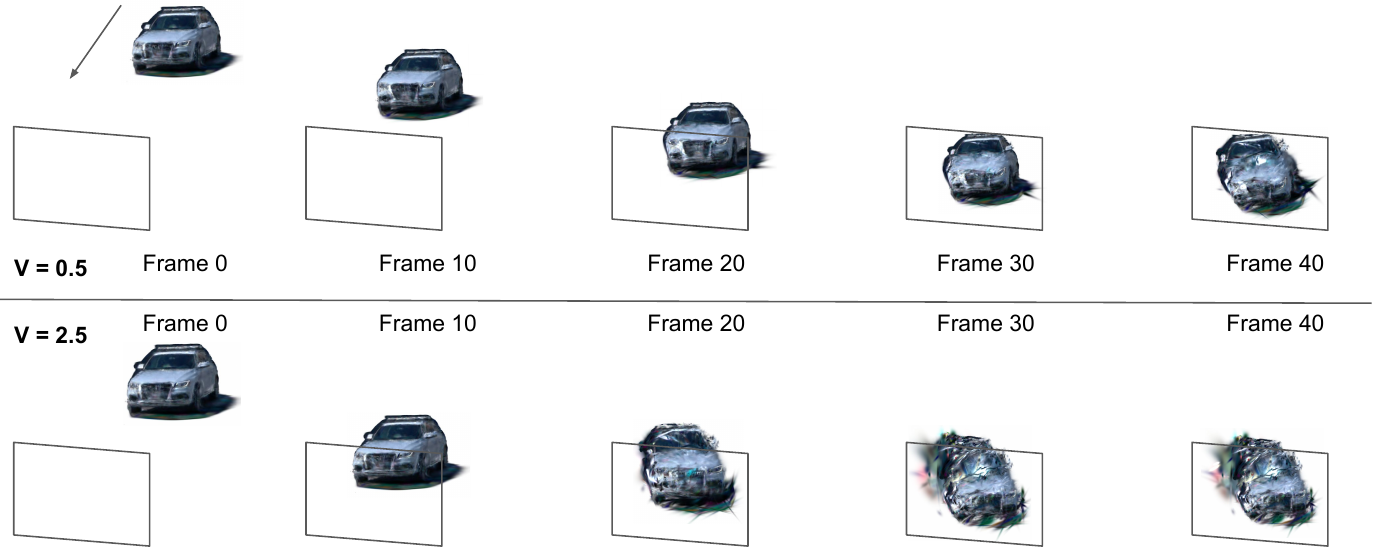}
  \caption{Simulation of a single-vehicle collision under different initial speeds. At 0.5 m/s, the vehicle hits the preset wall at frame 30 and exhibits mild frontal deformation, whereas at 2.5 m/s it shows more severe deformation.}
  \label{fig:4}
\end{figure*}

\begin{figure*}[t]
  \centering
  \includegraphics[width=0.9\textwidth]{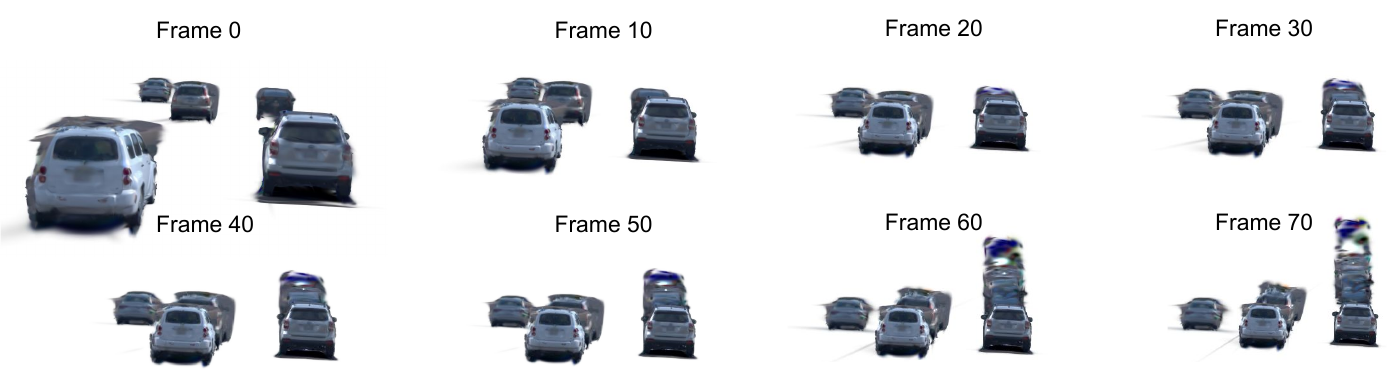}
  \caption{Simulation of a multi-vehicle collision. The two white SUVs closest to the camera are forced to collide with the stationary vehicles ahead.}
  \label{fig:5}
\end{figure*}

\subsubsection{Single Vehicle Collision}
Fig. \ref{fig:4} presents collision simulations at two different initial speeds. First, we reconstruct a white SUV from Sequence 031 and place a virtual collision plane directly in its path, assigning the vehicle a specified initial velocity. In the first row, the simulation uses $v_0 = 0.5\,\mathrm{m/s}$. Upon impact, kinetic energy is rapidly released, and the vehicle’s front deforms noticeably. In the second row, the initial velocity is increased to $v_0 = 2.5\,\mathrm{m/s}$, producing a faster collision and more severe deformation. Although these speeds are moderate in real-world terms, the deformations appear exaggerated because we have manually configured the vehicle’s material parameters to be more compliant, thereby emphasizing the collision response. Furthermore, due to limited access to accurate physical properties of real vehicles, the simulated behavior cannot exactly replicate reality. The physical parameters used in this collision test and in the two subsequent experiments are listed in Table \ref{Physical}.

\begin{table*}[!ht]
	\caption{Physical parameter settings for single-vehicle collision, multi-vehicle collision, and fall simulation.}
  \label{Physical}
	\begin{center}{%
		\begin{tabular}{c|cccc}
                \toprule
			Parameters & Single Vehicle Collision & Multi-vehicle Collision & Fall Simulation & Explanation\\
                \midrule
                $g (m/s)$ & 9.8 & 0 & 9.8 & Gravity \\
                $\nu$ & 0.4 & 0.4 & 0.4 & Poisson's Ratio \\
                $E (Pa)$ & 2e6 & 2e6 & 2e6 & Young's Modulus \\
                $\rho (kg/m^3)$ & 200 & 200 & 200 & Material Density \\
                $Grid$ & 50 & 200 & 200 & Grid Resolution\\
                $Material$ & Jelly & Jelly & Jelly & Material Type \\  
                \bottomrule
		\end{tabular}}
	\end{center}
\end{table*}

\subsubsection{Multi-vehicle Collision}
Fig. \ref{fig:5} shows collision simulations of multiple vehicles extracted from Sequence 002. The two white SUVs closest to the camera are assigned an initial velocity of $v_0 = 3.5\, \mathrm{m/s}$, while the remaining vehicles remain stationary. In this experiment, no virtual ground plane is used. Instead, we set $g = 0$ to keep all vehicles on the same horizontal plane. The right-hand SUV collides with the vehicle ahead around frame 20 and lifts it into midair. The left-hand SUV collides with the lead vehicle around frame 30, briefly pushing it forward before the lead vehicle moves laterally.

This experiment and the single-vehicle collision test presented earlier demonstrate both object-plane and object-object collisions. Such high-degree-of-freedom editing and physical simulation can fill gaps in the corner cases of autonomous driving datasets, which are less common and would lead to long-tailed training distributions\cite{zhang2023deep}. Moreover, collecting accident-scene data purely through real-world testing incurs high cost and safety risks. Therefore, by combining Gaussian scene editing with physics simulation, we can efficiently generate rich core-case data in the laboratory setting.

\begin{figure*}[t]
  \centering
  \includegraphics[width=1.0\textwidth]{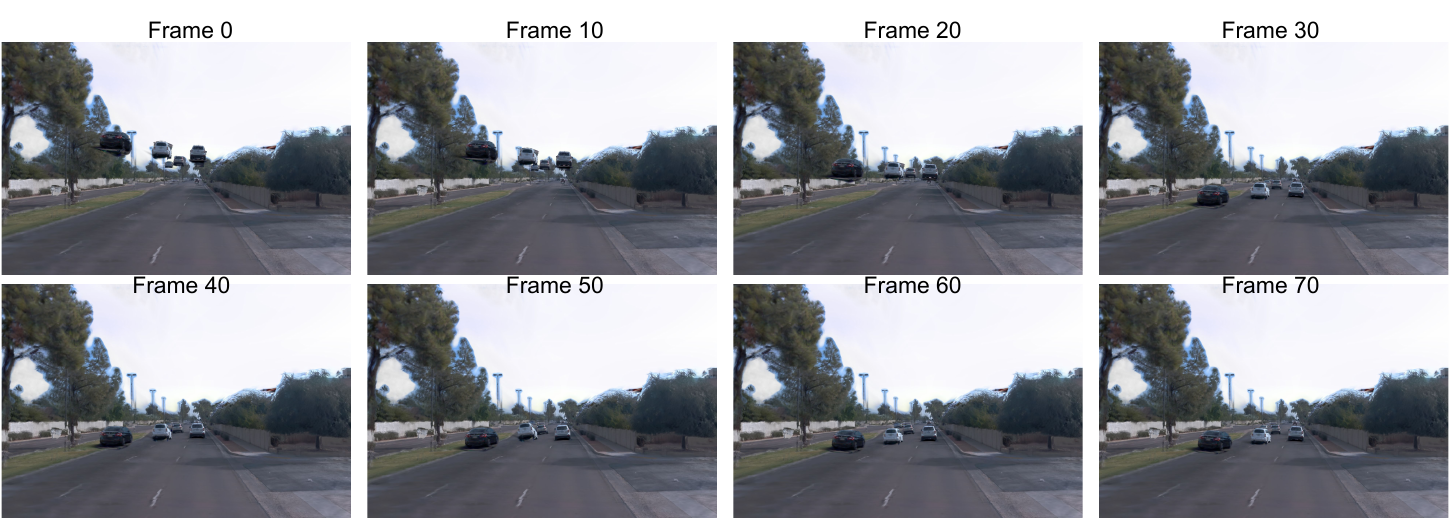}
  \caption{all simulation. The vehicles are first elevated to a predefined height and then released to freely fall onto the extracted ground plane.}
  \label{fig:7}
\end{figure*}

\subsubsection{Fall Simulation}
\begin{figure}[!ht]
  \centering
  \includegraphics[width=1.0\columnwidth]{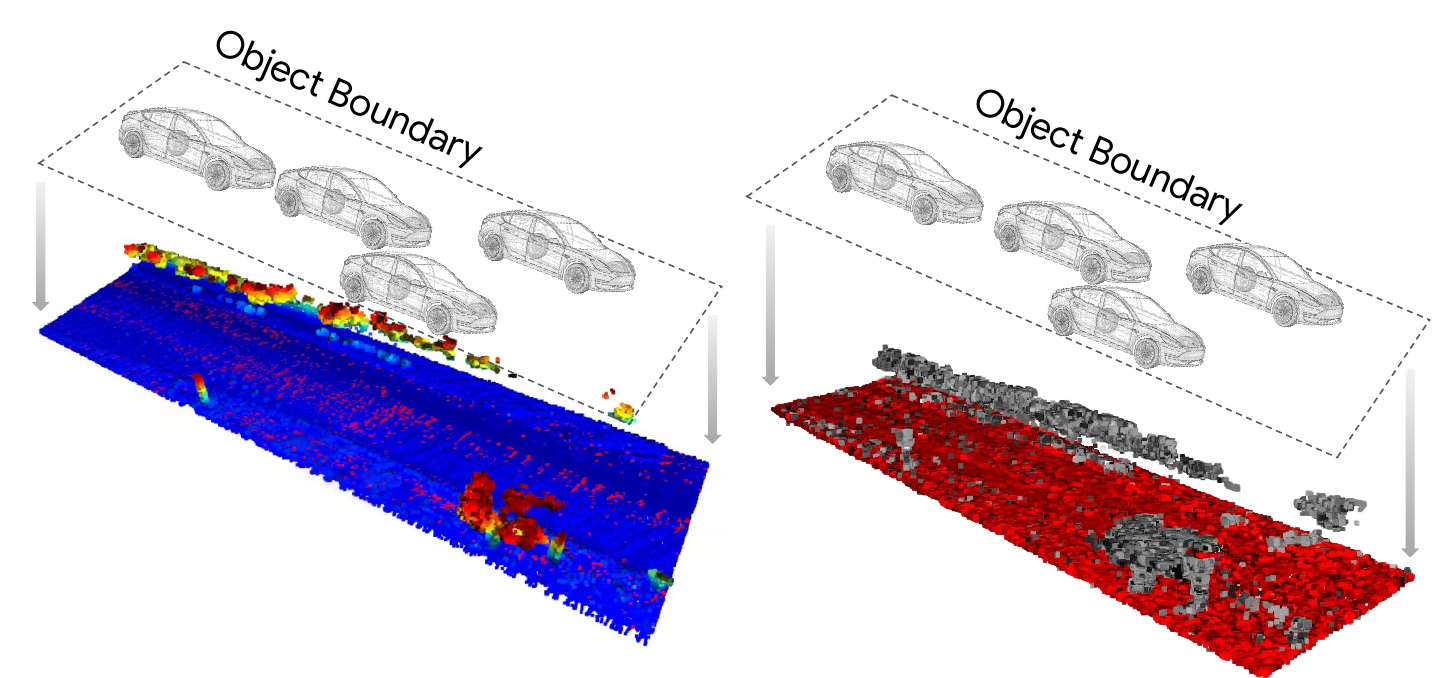}
  \caption{Road-plane detection using RANSAC (inlier points shown in red). The left image uses LiDAR inputs, while the right uses Gaussian points.}
  \label{fig:6}
\end{figure}

The previous two experiments present interactions between objects. This subsection simulates object-environment interaction by illustrating a vehicle, initially placed in midair, falling to the ground under gravity. We select Sequence 002 and first extract the road surface from the scene to provide a collision plane for the falling vehicle. Fig. \ref{fig:6} shows two methods for ground‐plane extraction: the left image uses LiDAR point clouds, and the right uses Gaussian point clouds. We begin by extracting the bounding boxes of all objects, projecting them vertically onto the background point cloud, and then filtering the points in the vicinity of each projected area. The RANSAC algorithm \cite{fischler1981random} is applied to these points to estimate the ground plane, and a surface collider is created to represent the road.

Fig. \ref{fig:7} illustrates the complete fall simulation. The vehicle begins to descend under gravity from frame 0, contacts the ground after approximately 0.5s, and following a slight bounce comes to rest on the road surface.

\section{Conclusion}
This paper introduces Real2Sim, an autonomous-driving scene simulation system built on Gaussian Splatting with full support for physics-level modeling. By representing each object as a cluster of Gaussian primitives, Real2Sim enables out-of-the-box high-degree-of-freedom scene editing. It further treats each Gaussian primitive as a particle and feeds it into an MPM-based physics engine, achieving realistic object-object and object-environment interactions. This unified framework provides an efficient, laboratory-scale approach for generating corner-case data, thereby supporting downstream tasks such as detection, tracking, motion prediction, and end-to-end autonomous-driving model training.

This work has several limitations that require further study. First, Gaussian modeling depends on the number of training viewpoints. In our editing and physics simulation experiments, some vehicles appear blurred or partially incomplete from unseen angles due to limited view coverage. Future work will address this by expanding the dataset or using diffusion models to refine and complete Gaussian representations. Second, rendering and physics simulation are currently decoupled. The simulation uses a Gaussian snapshot from a single timestep to predict subsequent physical behavior, which may deviate from the original Gaussian distribution and degrade rendering quality. Future work will explore an end-to-end joint estimation framework.
Finally, vehicle physical parameters require further calibration. Currently, each vehicle is modeled as a single soft body, limiting the simulation of detailed dynamics such as tire-road friction. We plan to develop decomposed Gaussian models that better reflect vehicle structures.
	
	\bibliographystyle{IEEEtran}
	\bibliography{root} 
	
\end{document}